\documentclass{article}
\usepackage{spconf,amsmath,graphicx}

\usepackage{epsfig} 
\usepackage{times} 
\usepackage{amsmath} 
\usepackage{amssymb}  

\title{Recognizing Focal Liver Lesions in Contrast-Enhanced Ultrasound with Discriminatively Trained Spatio-Temporal Model}
\name{Xiaodan Liang$^{\star}$ \qquad Qingxing Cao$^{\star}$ \qquad Rui Huang$^{\dagger}$ \qquad Liang Lin $^{\star}$\thanks{\small This work was supported by the Program of Guangzhou Zhujiang Star of Science and Technology (no. 2013J2200067), the Special Project on the Integration of Industry, Education and Research of Guangdong Province (no. 2012B091000101), the Guangdong Science and Technology Program (no. 2012B031500006).}}

\address{$^{\star}$ Sun Yat-sen University \qquad $^{\dagger}$ NEC Laboratories, China}

\begin{document}
%
\maketitle
\vspace{-4mm}
\begin{abstract}
\vspace{-2mm}

The aim of this study is to provide an automatic computational framework to assist clinicians in diagnosing Focal Liver Lesions (FLLs) in Contrast-Enhancement Ultrasound (CEUS). We represent FLLs in a CEUS video clip as an ensemble of Region-of-Interests (ROIs), whose locations are modeled as latent variables in a discriminative model. Different types of FLLs are characterized by both spatial and temporal enhancement patterns of the ROIs. The model is learned by iteratively inferring the optimal ROI locations and optimizing the model parameters. To efficiently search the optimal spatial and temporal locations of the ROIs, we propose a data-driven inference algorithm by combining effective spatial and temporal pruning. The experiments show that our method achieves promising results on the largest dataset in the literature (to the best of our knowledge), which we have made publicly available.
\end{abstract}
\vspace{-2mm}
%
\vspace{-4mm}
\section{Introduction}
\vspace{-4mm}

Liver cancer is the third cause of cancer-related death \cite{Guidelines12}. Visualization of Focal Liver Lesions (FLLs) has been attempted by employing various imaging techniques. Ultrasound is often performed in the diagnostics due to its low cost, efficiency and non-invasiveness. The use of Contrast-Enhanced Ultrasound (CEUS) can further assess the contrast enhancement (i.e., the intensity of the FLL area relative to that of the adjacent parenchyma) patterns of FLLs, which has markedly improved the accurate diagnosis of FLLs \cite{Guidelines12}. As shown in Fig.\ref{fig:tumor}, temporal enhancement patterns typically characterize the benign or malignant FLLs (e.g., sustain enhancement in the last two vascular phases for benign and hypo-enhancement for malignant FLLs). On the other hand, spatial enhancement patterns during the arterial phase often characterize the specific types of FLLs.

Extensive research efforts have been made to assist the experts in diagnosing different types of cancers and, in particular, FLLs using ultrasound images \cite{tissueChar}\cite{ROIselectionFLL13}. However, the application of CEUS for differentiating FLLs is still a relatively new technique \cite{CEUSANN08}\cite{CEUSDVP11}\cite{CEUSmotion12}\cite{BakasSpotFrame13}.  A cascade of Artificial Neural Networks\cite{CEUSANN08} is employed to classify FLLs based on manually segmented lesion regions. Anaye et al. \cite{CEUSDVP11} analyzes the Dynamic Vascular Patterns (DVPs) of FLLs with respect to surrounding healthy parenchyma to differentiate between benign and malignant FLLs. In \cite{CEUSmotion12}, Bakas et al. track a manually initialized FLL and its surrounding parenchyma to characterize it as either benign or malignant based on its vascular signature. In their recent work \cite{BakasSpotFrame13}, an automated method for selection of the optimal frame for initialization of the FLL candidates is proposed.


\begin{small}
\begin{figure}[ptb]
\begin{center}
\epsfig{file=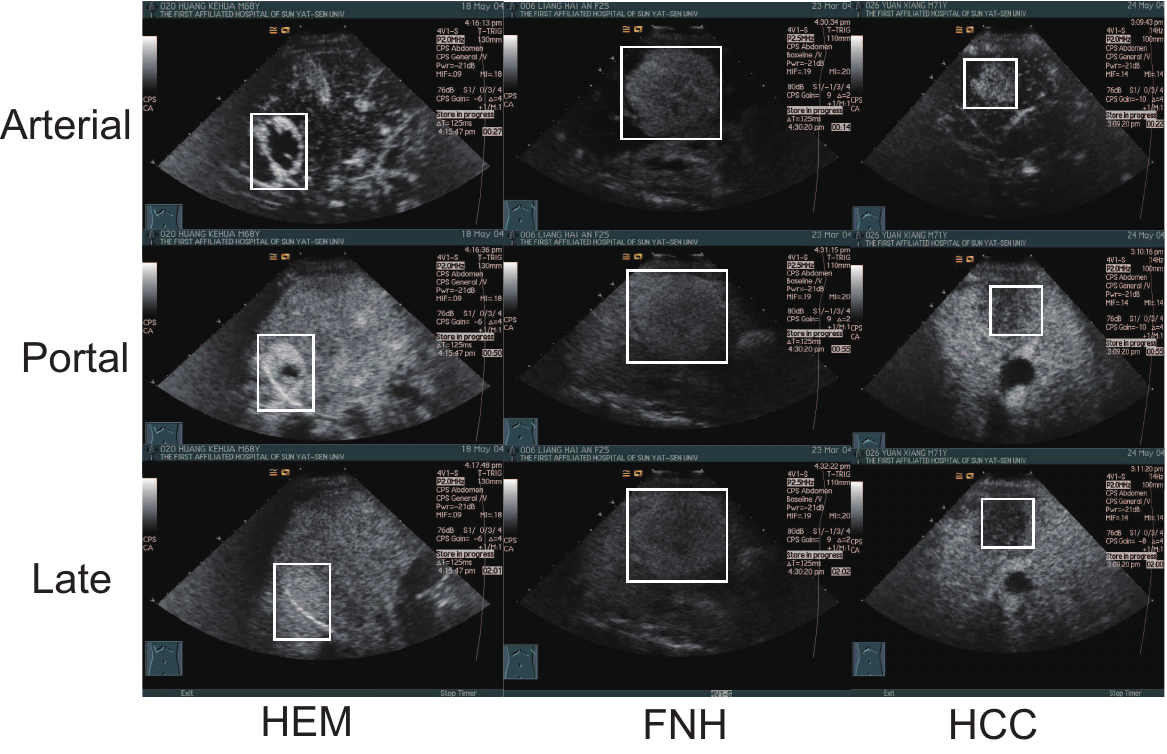, width=3in}
\end{center}
\vspace{-7mm}
\caption{The enhancement pattern ROIs of three different FLLs: Hemangioma (HEM), Focal Nodular Hyperplasia (FNH), Hepatocellular Carcinoma (HCC), in three different phases: the arterial, portal venous and late phases. The HEM and FNH are benign FLLs and HCC is a malignant FLL.}
\vspace{-6mm}
\label{fig:tumor}
\end{figure}
\end{small}

In all these works, varying degrees of manual interactions are required to identify the Regions of Interest (ROIs) of FLLs or the normal parenchyma. The manual annotations are highly dependent on the skills and knowledge of the experts, leading to large variations in inter-/intra-observer image interpretations. Besides, the ever-increasing amount of CEUS data acquired and processed nowadays demands automatic computational systems that can save the radiologists' time and efforts. In addition, most of the previous works focused on differentiating between benign and malignant FLLs, or characterizing a specific type of FLLs. We, on the other hand, are trying to combine different enhancement patterns to recognize multiple different types of FLLs in a unified framework.

The main contributions of our work herein are threefold. First, we propose a fully automatic computational framework to recognize FLLs by modeling the locations of ROIs as latent variables in a discriminative model and combining both spatial and temporal enhancement patterns of the ROIs into the framework. Our model is then trained by a weakly supervised learning algorithm, which alternates between inferring the most probable spatial and temporal locations of the ROIs and optimizing the model parameters. Second, considering that most of the video frames and the regions in each frame contain redundant or irrelevant information for recognizing FLLs, the automatic detection of optimal locations of the ROIs is made very efficient by a novel data-driven inference method, which combines the spatial and temporal pruning techniques to disregard less discriminative frames and regions. The optimal ROI locations are then determined by dynamic programming. Last but not least, a new region representation for ROIs is presented to capture the important and relevant ultrasonic characteristics of FLLs, which is not necessarily limited to our framework.

We apply our method on a new dataset (namely SYSU-CEUS dataset) we collected and made public, which contains in total 353 CEUS video sequences of three types of FLLs ($186$ HCC, $109$ HEM and $58$ FNH), and is, to the best of our knowledge, the largest dataset in the literature. The experimental results demonstrate that our method achieves promising performance without manual interactions.



\vspace{-5mm}
\section{Our Model}
\label{sec:model}
\vspace{-4mm}

\subsection{Region representation}
\label{sec:feature}
\vspace{-2mm}

The accurate classification of FLLs highly depends on the representation of the characteristics of the lesion regions (e.g., internal echo, morphology, edge, echogenicity and posterior echo enhancement). However, one single ROI $R$ is often insufficient to capture all the ultrasonic characteristics. For instance, the region inside the lesion, denoted as $R^-$, can capture the internal echo of the FLL; the lesion region $R$ can be used to observe the boundary and the morphology of the FLL; and the tissue area surrounding the lesions, denoted as $R^+$, can be used to measure the posterior echo enhancement. The echogenicity of the lesion can be measured by comparing the intensities of above regions. Thus, given an ROI $R$, the regions $R^-$ and $R^+$ can be obtained by shrinking and enlarging $R$ by a small factor, respectively. We then propose an effective region representation as following:

\vspace{-6mm}
{\small
\begin{equation}\begin{aligned}
f(R)=[f^t(R^-), f^t(R), f^t(R^+), f^d(R^-,R), f^d(R, R^+)]
\end{aligned}\end{equation}
}\noindent
where $f^t$ extracts the appearance features of each region, such as Grey Level Co-occurrence Matrix(GLCM) and Local Phase(LP); $f^d$ calculates the mean intensity difference of two regions. Consequently, the concatenation of all these features, $f(R)$, captures all the desired ultrasonic characteristics of this region $R$.

\vspace{-5mm}
\subsection{Model representation}
\vspace{-2mm}

Given a CEUS video sequence $\mathbf{x}$, $y$ is the corresponding class label of the FLL in this video, ranging over a finite set $\mathcal{Y}$ (e.g., $\mathcal{Y}$=\{HCC, HEM, FNH\}). We assume that the FLL can be compactly represented by a set of ROIs $\{R_1,R_2,\dots,R_m\}$ in three vascular phases: arterial, portal venous, and late phases. Intuitively, ROIs are the most discriminative regions for distinguishing different FLLs. And each ROI $R_i$ is a region extracted from the video frame $t_i$, at the spatial location $p_i = (x_i, y_i, s_i)$, where $x_i, y_i, s_i$ are the coordinates and the scale of the ROI. The latent variables $\mathbf{h} = \{h_1, h_2, \dots, h_m\}$, where $h_i = (p_i, t_i)$, is the location of $R_i$, taken values from a finite set $\mathcal{H}_i$ of all possible ROI locations. Given video $\mathbf{x}$, its corresponding class label $y$, and latent variables $\mathbf{h}$, the conditional probability of the recognition problem is defined as,

\vspace{-6mm}
{\small
\begin{equation}\begin{aligned}
    p(y|\mathbf{x};\omega) &= \sum_{\mathbf{h}\in\mathcal{H}}p(y,\mathbf{h}|\mathbf{x};\omega)\\ &=\frac{\sum_{\mathbf{h}\in\mathcal{H}}\exp(\omega^\mathrm{T}\cdot\psi(\mathbf{x},\mathbf{h},y))}{\sum_{\hat{y}\in\mathcal{Y}}\sum_{\mathbf{h}\in\mathcal{H}}\exp(\omega^\mathrm{T}\cdot\psi(\mathbf{x},\mathbf{h},\hat{y}))}
    \label{eq:logeq}
\end{aligned}\end{equation}
}\noindent
where $\omega$ is the model parameter vector, {\small $\mathcal{H}=\mathcal{H}_1\times\mathcal{H}_2\times\cdots\times\mathcal{H}_m$}, and $\psi(\mathbf{x},\mathbf{h},y)$ is a feature vector depending on the video sequence $\mathbf{x}$, the class label $y$, and the latent variables $\mathbf{h}$. We define the formulation of $\omega^\mathrm{T}\cdot\psi(\mathbf{x},\mathbf{h},y)$ as the following, including two terms: unary potential and pairwise potential,

\vspace{-4mm}
{\small
\begin{equation}\begin{aligned}
    \omega^\mathrm{T}\cdot\psi(\mathbf{x},\mathbf{h},y) &= \sum_{i\in m}\alpha_i^\mathrm{T}\cdot\phi^u(\mathbf{x},y,h_i)\\
    &+ \sum_{(i,j)\in\mathcal{E}}\beta_{i,j}^\mathrm{T}\cdot\phi^p(\mathbf{x},y,h_i,h_{j})
    \label{eq:global}
\end{aligned}\end{equation}
}\noindent
where $\phi^u(\cdot)$ is the unary potential function of variable $h_i$ and $\phi^p(\cdot)$ is the pairwise potential function of $(h_i,h_{j})$. $\mathcal{E}$ is the set of neighboring latent variables (defined for the pairs of temporally adjacent ROIs).

1) Unary potential $\alpha_i^\mathrm{T}\cdot\phi^u(\mathbf{x},y,h_i)$: This singleton potential function $\phi^u(\cdot)$ models the compatibility between class label $y$ and appearance of region $R_i$ (note that $R_i = \mathbf{x}(h_i)$).

\vspace{-5mm}
{\small
\begin{equation}\begin{aligned}
\alpha_i^\mathrm{T}\cdot\phi^u(\mathbf{x},y,h_i) = \sum_{a\in\mathcal{Y}}\sum_{b\in\mathcal{H}_i}\alpha_{i}^a \cdot \delta_y(a)\cdot\delta_{h_i}(b)\cdot f(\mathbf{x}(h_i))
\label{eq:unary}
\end{aligned}\end{equation}
}\noindent
where $f(\mathbf{x}(h_i))$ is the feature vector describing the appearance of the region, as defined in section \ref{sec:feature}. The indicator function $\delta_y(a)$ is equal to one if $y=a$, zero otherwise. Similarly, $\delta_{h_i}(b)$ is equal to one if $h_i=b$, zero otherwise. The parameter $\alpha_i$ is simply the concatenation of all $\alpha_{i}^a$.

2) Pairwise potential $\beta_{i,j}^\mathrm{T}\cdot\phi^p(\mathbf{x},y,h_i,h_{j})$: The potential function $\phi^p(\cdot)$ models the compatibility between class label $y$ and the temporal transition of a pair of neighboring latent variables $(h_i,h_{j})$.

\vspace{-3mm}
{\small
\begin{equation}\begin{aligned}
&\beta_{i,j}^\mathrm{T}\cdot\phi^p(\mathbf{x},y,h_i,h_{j}) = \sum_{a\in\mathcal{Y}}\sum_{b\in\mathcal{H}_i}\sum_{c\in\mathcal{H}_{j}}\\
&\beta_{i,j}^a \cdot \delta_y(a)\cdot\delta_{h_i}(b)\cdot\delta_{h_j}(c)\cdot f^p(\mathbf{x},h_i,h_{j})
\label{eq:pairwise}
\end{aligned}\end{equation}
}\noindent
where $f^p(\cdot)$ includes two components: appearance variance feature, computed by the difference of $f(\mathbf{x}(h_i))$ and $f(\mathbf{x}(h_j))$, and spatial displacement feature, i.e., Euclidean distance between the spatial coordinates of $h_i$ and $h_j$. And the parameter $\beta_{i,j}$ is simply the concatenation of all $\beta_{i,j}^a$.

\vspace{-5mm}
\subsection{Learning}
\label{sec:learning}
\vspace{-3mm}

Given a training set $D = \{(\mathbf{x}_1,y_1), \dots, (\mathbf{x}_n,y_n)\}$, the model parameter $\omega$ can be learned by maximizing the conditional log-likelihood on the training samples:

\vspace{-7mm}
{\small
\begin{equation}\begin{aligned}
    \omega^* &= \arg \max_{\omega}\mathcal{L}(\omega) = \arg \max_{\omega}\sum_{i=1}^N\mathcal{L}^i(\omega)\\
    &= \arg \max_{\omega}\sum_{i=1}^N \log p(y_i|\mathbf{x}_i;\omega)\\
    &= \arg \max_{\omega}\sum_{i=1}^N \log(\sum_{\mathbf{h}\in \mathcal{H}}p(y_i,\mathbf{h}|\mathbf{x}_i;\omega))
\end{aligned}\end{equation}
}\noindent
where $\mathcal{L}^i(\omega)$ denotes the conditional log-likelihood of the $i^{th}$ training example, defined in Eq(\ref{eq:logeq}), and $\mathcal{L}(\omega)$ denotes the conditional log-likelihood of the whole training set. The objective function $\mathcal{L}(\omega)$ is not concave, due to the latent variables $\mathbf{h}$. We adopt the latent structural SVM learning framework \cite{LSVM}, which alternates between inferring the latent variables $\mathbf{h}$ and optimizing the model parameter $\omega$. The problem of inferring $\mathbf{h}$ can be solved efficiently using a data-driven inference algorithm (Sec. \ref{sec:inference}), and the parameter optimization is a standard structural SVM training problem, solved by the cutting-plane algorithm.  We repeat the above two steps until convergence. We use the one-vs-one binary classification strategy for multi-class classification problem.

Given a learned model, the classification is achieved by first finding the best hypothesis $\{h_i\}_1^m$ for $m$ ROIs, then picking the FLL class with the highest SVM score. The score of an example $\mathbf{x}$ with a learned classifier is defined as:

\vspace{-4mm}
\begin{equation}\begin{aligned}
    f_\omega(\mathbf{x},y) = \max_{\mathbf{h}\in \mathcal{H}} \omega^\mathrm{T} \cdot \psi(\mathbf{x},y,\mathbf{h})
    \label{equ:op}
\end{aligned}\end{equation}

\vspace{-7mm}
\subsection{Data-driven inference}
\label{sec:inference}
\vspace{-3mm}
The inference task is to find the optimal locations of the ROIs (i.e., the latent variables $h$). However, the searching space will be very large if we consider all regions in all frames. Thus, we propose a data-driven inference algorithm, which efficiently combines the spatial and temporal pruning techniques to disregard less discriminative frames and regions. The optimal locations $\{h_i\}_1^m$ of the most discriminative ROIs can then be determined using dynamic programming.

1) Temporal pruning: In a CEUS video, the appearance of ultrasound frames often varies slowly and smoothly according to the hemodynamic, and the most discriminative frames are usually those with the largest contrast changes compared with neighboring frames. Thus, a small set of candidate frames, which have local maximum of the contrast change, are automatically selected. In particular, for each frame $I_t, (t=1,\cdots,T)$ in a video $\mathbf{x}$, we compute the contrast feature $v_t$ from the co-occurrence distribution $C_t$ defined over $I_t$ \cite{FeatureTexture73}. The contrast vector $\mathbf{v}$ is then $(v_1,v_2,\dots,v_T)$. Let $\Delta\mathbf{v}$ be the gradient of $\mathbf{v}$, the candidate frame set $B$ is formed by finding the frames at the local maximum of $\Delta\mathbf{v}$.

2) Spatial pruning: After temporal pruning, we also prune the less important regions by considering two priors: saliency prior and location prior. First, we believe that {\it salient} regions (e.g., having higher contrast or containing typical structures) have more discriminative information, and thus are more likely to be candidates of ROIs. Second, we observe that FLLs often appear in or close to the center of the images, probably because a skilled ultrasound operator usually places the liver area in the middle of the display. According to these two observations, we evaluate all the regions with different scales in each candidate frame $I\in B$ (sliding window protocol), and only select the regions with prior probability larger than a threshold $\tau$ as ROI candidates. The prior probability of a region $r$ being an ROI is,

\vspace{-5mm}
{\small
\begin{equation}\begin{aligned}
    p(r) = \mathcal{S}(r)\mathcal{G}(C^r|C^I, \sigma)
 \end{aligned}\end{equation}
}\noindent
where $\mathcal{S}(r)$ is the normalized mean saliency of the region $r$ in the saliency map $\mathcal{S}$, computed by the quaternion-based spectral saliency method \cite{Saliency} on image $I$. $C^r$ and $C^I$ are the centroid of region $r$ and the image $I$, respectively.  $\mathcal{G}(C^r|C^I,\sigma)$ is a Gaussian distribution.

It is worth noting that the spatial pruning in the last two vascular phases (portal and late) can be more aggressive. This is because the contrast between FLLs and normal tissues is often very low, and the locations of FLLs do not change much since the arterial phase. Thus, in the last two phases, we only search the regions in a spatial neighborhood around the locations of ROI candidates found in the arterial phase. Finally, given the model parameters and the observations, the latent variables $\mathbf{h} = \{h_1,h_2,\dots,h_m\}$ form a hidden Markov model, and can be solved exactly by the Viterbi algorithm \cite{koller2009probabilistic}.

\begin{small}
\begin{table}
  \centering
  \begin{tabular}{| c | c | c | c | c |}
    \hline
     & Sens$^{Benign}$
     & Sens$^{Malignant}$
     & Accuracy\\
     \hline
    Ours & \textbf{85.7\%} & \textbf{93.4\%} & \textbf{89.7\%} \\
    \hline
  \end{tabular}
  \vspace{-3mm}
  \caption{Sensitivities and mean accuracies on characterizing benign and malignant FLLs. Sens means the sensitivity of the specific class.}
  \vspace{-4mm}
  \label{Tab:comparison}
\end{table}
\end{small}

\begin{small}
\begin{table}
  \begin{tabular}{| c | c | c | c | c |}
    \hline
     &Sens$^{HCC}$
     & Sens$^{HEM}$
     & Sens$^{FNH}$
     & Acc\\
    \hline
    DDI & \textbf{88.9\%} & 81.0\% & 63.6\% & \textbf{82.4\%} \\
    manual & 86.1\% & \textbf{85.7\%} & \textbf{72.7\%} & \textbf{83.8\%} \\
    bruteforce & 83.3\% & 80.1\% & 36.4\% & 75.0\% \\
    baseline &78.9\% & 22.0\% & 10.3\% & 49.9\% \\
    \hline
  \end{tabular}
  \vspace{-4mm}
  \caption{Sensitivities and mean accuracies in the different experiment settings.}
  \vspace{-7mm}
  \label{Tab:inference}
\end{table}
\end{small}

\vspace{-5mm}
\section{Results}
\label{sec:result}
\vspace{-4mm}


We test our method on the SYSU-CEUS dataset collected from the First Affiliated Hospital, Sun Yat-sen University, which is public available{\small\footnote{https://github.com/lemondan/Focal-liver-lesions-dataset-in-CEUS}}. The equipment used was Aplio SSA-770A (Toshiba Medical System). The dataset consists of three types of FLLs: $186$ HCC, $109$ HEM and $58$ FNH instances (i.e., 186 malignant and 167 benign instances). All these instances with resolution $768*576$ were taken from different patients, with large variations in appearance and enhancement patterns (e.g., size, contrast, shape and location) of FLLs. We adopt the 5-fold cross validation training strategy and the sensitivity for each class and mean accuracy as the evaluation criteria, similar to \cite{CEUSDVP11}. In our implementation, we extract four statistics (i.e., Contrast, Correlation, Energy, Homogeneity) of GLCM \cite{FeatureTexture73} with four orientations ($\theta = 0^o,45^o,90^o,135^o$), and one distance ``1'', to represent the texture feature $f^t$ (\ref{sec:feature}). Three scales of regions (i.e., $64\times 64$, $128\times 128$, $200\times 200$) and step length 20 are used for sliding windows, and $\tau = 0.6$ and $\sigma = 0.5$ are used for spatial pruning. The experiments are carried out on a PC with Core I7 3.4GHz CPU, and the average processing time for a 4-min CEUS video is about 100 seconds.

\begin{small}
\begin{table}
  \begin{tabular}{| c | c | c | c | c | c | c | c | }
    \hline
     & Sens$^{HCC}$
     & Sens$^{HEM}$
     & Sens$^{FNH}$
     & ACC1\\
    \hline
    \cite{ROIselectionFLL13}$^{GLCM}$  &85.9 \% & 75.9\% & 36.2\% & 74.7\% \\
    \cite{ROIFHL09}$^{GLCM}$ & \textbf{88.1\%} & 67.5\% & 51.7\% & 75.8\%\\
    \cite{ROIxian10}$^{GLCM}$ & 82.1\% & 61.1\% & 34.4\% & 67.8\%\\

    Ours$^{GLCM}$ & 87.2\% & \textbf{83.5\%} & \textbf{67.2\%} & \textbf{82.7\%}\\
    \hline

    \cite{ROIselectionFLL13}$^{Law's}$  &82.7 \% & 75.9\% & \textbf{72.4\%} & 78.9\%\\
    \cite{ROIFHL09}$^{Law's}$ & 75.6\% & 77.7\% & 62.0\% & 74.2\%\\
    \cite{ROIxian10}$^{Law's}$ & 69.7\% & 72.2\% & 56.9\% & 68.4\%\\

    Ours$^{Law's}$ & \textbf{84.3\%} & \textbf{87.2\%} & 67.6\% & \textbf{82.4\%}\\
    \hline
    \cite{ROIselectionFLL13}$^{LP}$  &85.9 \% & 67.5\% & 63.7\% & 76.5\%\\
    \cite{ROIFHL09}$^{LP}$ & 80.0\% & 69.4\% & 55.1\% & 72.7\%\\
    \cite{ROIxian10}$^{LP}$ & 78.9\% & 50.9\% & 48.2\% & 65.2\%\\
    Ours$^{LP}$ & \textbf{86.1\%} & \textbf{73.4\%} & \textbf{63.8\%} & \textbf{78.3\%}\\
    \hline

  \end{tabular}
  \vspace{-3mm}
  \caption{Comparisons of region representation methods by applying different feature descriptors.}
  \vspace{-4mm}
  \label{Tab:Feature}
\end{table}
\end{small}

We first report the sensitivities and mean accuracies of our method in differentiating benign and malignant FLLs in Table. \ref{Tab:comparison}. The average accuracy ($89.7\%$) is comparable, if not superior, to the results reported in previous studies on smaller datasets \cite{CEUSDVP11}\cite{CEUSmotion12}.The second experiment in Table. \ref{Tab:inference} shows the effectiveness of our data-driven inference algorithm by altering the procedure to determine the ROIs. Our data-driven inference algorithm (``DDI") is compared with 1) ``manual": the ROI of each instance in the arterial phase is manually selected and the inference only performed in the portal and late phase; 2) ``bruteforce": the liver region is labeled and the optimal ROIs are searched in the entire region of liver, without pruning; 3) ``baseline": the ROIs are randomly selected in the images of three phases. The results demonstrate that our fully automatic inference algorithm achieves comparable performance to the ``manual'' method, and performs better than ``brute force'' and ``baseline''. Note that the performance of our algorithm on FNH is worse because the amount of training data of FNH is relatively small.

Finally, in Table.\ref{Tab:Feature} we compare the region representation of our framework with other state-of-the-art methods: Multiple-ROI \cite{ROIselectionFLL13}, ROI$^{posterior}$ \cite{ROIFHL09} and ROI$^{out}$ \cite{ROIxian10}. Each region representation is tested with three popular low-level features: GLCM, Law's texture, and Local Phase, similar to \cite{ROIselectionFLL13}. We manually select ROIs in three phases as required in previous works (note here we do not consider the performance of the inference algorithm), and use linear SVM as the classifier. The results show that our region representation obtains superior performances in general.


\vspace{-6mm}
\section{CONCLUSIONS}
\label{sec:conclusion}
\vspace{-4mm}
In this work we propose a fully automatic computational framework for characterizing different types of FLLs in CEUS, which efficiently combines the diverse information of spatial and temporal enhancement patterns. Besides, a weakly supervised learning algorithm is utilized, which alternates between inferring the latent variables (i.e. the locations of ROIs) and optimizing the model parameters. An efficient data-driven inference algorithm is then proposed to efficiently determine the optimal locations of ROIs. The results show promising classification accuracies and the potential of being developed for real-time clinical applications. In the future, a more interactive system will be developed to enable the radiologists to revise the diagnosis according to the detailed outputs of our algorithm (e.g., the locations of ROIs and the reference frames).

\vspace{-5mm}
\small{\bibliographystyle{IEEEbib}
\bibliography{strings,refs}

\begin{thebibliography}{10}

\bibitem{Guidelines12}
Michel Claudon and et~al.,
\newblock ``Guidelines and good clinical practice recommendations for contrast
  enhanced ultrasound in the liver -update 2012,''
\newblock {\em Ultrasound in Medicine and Biology}, vol. 39, no. 2, 2013.

\bibitem{tissueChar}
J.~Alison Noble,
\newblock ``Ultrasound image segmentation and tissue characterization,''
\newblock {\em Proc Inst Mech Eng H.}, vol. 224, no. 2, pp. 307--16, 2010.

\bibitem{ROIselectionFLL13}
Jae~Hyun Jeon and et~al.,
\newblock ``Multiple roi selection based focal liver lesion classification in
  ultrasound images,''
\newblock {\em Expert Systems with Applications}, vol. 40, no. 2, 2013.

\bibitem{CEUSANN08}
J~Shiraishi, K~Sugimoto, F~Moriyasu, N~Kamiyama, and K~Doi,
\newblock ``Computer-aided diagnosis for the classification of focal liver
  lesions by use of contrast-enhanced ultrasonography.,''
\newblock {\em Med Phys.}, vol. 35, no. 5, pp. 1734--46, 2008.

\bibitem{CEUSDVP11}
A~Anaye and et~al.,
\newblock ``Differentiation of focal liver lesions: usefulness of parametric
  imaging with contrast-enhanced us.,''
\newblock {\em Radiology}, vol. 26, no. 1, 2011.

\bibitem{CEUSmotion12}
Spyridon Bakas and et~al.,
\newblock ``Histogram-based motion segmentation and characterisation of focal
  liver lesions in ceus,''
\newblock {\em Annals of the BMVA.}, vol. 7, 2012.

\bibitem{BakasSpotFrame13}
Spyridon Bakas and et~al.,
\newblock ``Spot the best frame: Towards intelligent automated selection of the
  optimal frame for initialisation of focal liver lesion candidates in
  contrast-enhanced ultrasound video sequences,''
\newblock in {\em Intelligent Environments(IE)}, 2013.

\bibitem{LSVM}
Chun-Nam~John Yu and Thorsten Joachims,
\newblock ``Learning structural svms with latent variables,''
\newblock in {\em ICML}, 2009.

\bibitem{FeatureTexture73}
R~Haralick, K~Shanmugam, and Its\'Hak Dinstein,
\newblock ``Textural features for image classification,''
\newblock {\em Systems, Man and Cybernetics, IEEE Transactions on}, vol. SMC-3,
  no. 6, 1973.

\bibitem{Saliency}
Boris Schauerte and Rainer Stiefelhagen,
\newblock ``Quaternion-based spectral saliency detection for eye fixation
  prediction,''
\newblock in {\em ECCV}, 2012, pp. 116--129.

\bibitem{koller2009probabilistic}
Daphne Koller and Nir Friedman,
\newblock {\em Probabilistic Graphical Models - Principles and Techniques.},
\newblock MIT Press, 2009.

\bibitem{ROIFHL09}
SH~Kim and et~al.,
\newblock ``Computer-aided image analysis of focal hepatic lesions in
  ultrasonography: preliminary results.,''
\newblock {\em Abdom Imaging}, vol. 34, no. 2, 2009.

\bibitem{ROIxian10}
Guang ming Xian,
\newblock ``An identification method of malignant and benign liver tumors from
  ultrasonography based on glcm texture features and fuzzy svm,''
\newblock {\em Expert Systems with Applications}, vol. 37, no. 10, pp. 6737 --
  6741, 2010.

\end{thebibliography}
}

\end{document}